\documentclass{article}
\usepackage{spconf,amsmath,graphicx}
\usepackage{subcaption}  % For subfigures

\usepackage{enumitem}
\setlist{nosep, leftmargin=14pt}
\usepackage{textcomp}

\usepackage{graphicx}
\usepackage{amsmath}
\usepackage{amssymb}
\usepackage{booktabs}
\usepackage{xparse} % per definire comandi moderni

\usepackage{multirow}
\usepackage{hyperref}
\usepackage{url}
\usepackage{siunitx}
\usepackage{cite}

\usepackage{mwe} % to get dummy images
\usepackage{comment}
\newcommand{\mg}[1]{ \textcolor[rgb]{0.72,0.45,0.2}{{\bf MG: #1}}}
\usepackage{cleveref}  % in the preamble

\def\x{{\mathbf x}}

\title{Synthetic Dataset Generation and Validation \\ for Robotic Surgery Instrument Segmentation}
\name{\begin{tabular}{c} Giorgio Chiesa \quad Rossella Borra \quad Vittorio Lauro \quad Sabrina De Cillis \\ \quad Daniele Amparore \quad Cristian Fiori \quad Riccardo Renzulli \quad Marco Grangetto\end{tabular}}
\address{University of Turin, Italy }
\begin{document}
%\ninept
%
\maketitle
%}
\begin{abstract}
This paper presents a comprehensive workflow for generating and validating a synthetic dataset designed for robotic surgery instrument segmentation. A 3D reconstruction of the Da Vinci\texttrademark \ robotic arms was refined and animated in Autodesk Maya\texttrademark \ through a fully automated Python-based pipeline capable of producing photorealistic, labeled video sequences. Each scene integrates randomized motion patterns, lighting variations, and synthetic blood textures to mimic intraoperative variability while preserving pixel-accurate ground truth masks.
To validate the realism and effectiveness of the generated data, several segmentation models were trained under controlled ratios of real and synthetic data. Results demonstrate that a balanced composition of real and synthetic samples significantly improves model generalization compared to training on real data only, while excessive reliance on synthetic data introduces a measurable domain shift. The proposed framework provides a reproducible and scalable tool for surgical computer vision, supporting future research in data augmentation, domain adaptation, and simulation-based pretraining for robotic-assisted surgery. Data and code are available at \url{https://github.com/EIDOSLAB/Sintetic-dataset-DaVinci}
\end{abstract}
\section{Introduction}
\label{sec:intro}
Robotic surgery has transformed minimally invasive procedures by improving their precision, ergonomics and reproducibility. During the past ten years, robotic systems like Da Vinci\texttrademark \  surgical system have evolved from mechanical telemanipulators to intelligent platforms capable of helping surgeons in complex decision-making processes.
Instrument segmentation plays a critical role in enabling visual understanding during robotic surgery. Accurate identification of the position and boundaries of surgical instruments is essential for a large downstream applications. Inaccurate segmentation directly affects the performance of these systems, with the risk of reducing reliability or compromising safety in clinical applications. 

Deep learning-based solutions require large volumes of pixel-level annotations to achieve reliable segmentation and detection; however, obtaining such annotations in surgical settings is highly challenging.
The collection of real annotated datasets for robotic instrument segmentation remains a major bottleneck. Strict privacy regulations protect surgical videos, and their annotation requires domain expertise and significant manual effort. Additionally, intraoperative variability makes it challenging to achieve dataset generalization. Synthetic data generation has thus emerged as a promising strategy to overcome these limitations, providing scalable, controllable, and privacy-preserving alternatives to real data.

In this work, we present a comprehensive and reproducible workflow for generating and validating synthetic data for robotic surgery instrument segmentation. The proposed pipeline enables the creation of photorealistic surgical scenes through a fully automated Python–Maya\texttrademark \ framework.
Using this framework, we generated a high-fidelity synthetic dataset of Da Vinci\textsuperscript{TM} robotic tools with pixel-accurate ground truth masks. The dataset was validated experimentally by training segmentation networks with varying ratios of real and synthetic samples.

\section{3d model acquisition}
To reconstruct the 3D model of the surgical robot's instruments, we used photogrammetry with the following method.
A Canon EOS 2000D was used during the acquisition phase, configured to a resolution of 6000×4000 pixels, 24-bit color depth, an aperture of f/3.5, an exposure time of 1/60 s, an ISO of 200, and an 18 mm focal length. The imaging environment corresponded to a hospital room artificially illuminated with neon lighting, providing a correlated color temperature between 4000 K and 4500 K. To ensure metric accuracy were positioned two colored markers, two black-and-white markers, and four calibrated reference objects.
The robotic system, characterized by a dominant parallelepiped combined with irregular geometric features, was documented through a mixed structured trajectory consisting of two major spirals alternated with four linear trajectories and four secondary spirals. This acquisition pattern, aligned with the multiple acquisition principles outlined in PODS \cite{Lauro_Lombardo_2025}, guaranteed complete spatial coverage of critical regions—particularly metallic joints—achieving an average image overlap of approximately $70\%$. A total of 597 photographs were captured over approximately 27 minutes under stable distance conditions, with temporary shielding used to prevent direct reflections. Image alignment and Structure-from-Motion reconstruction were conducted using Zephyr 8.011 \cite{zephyr_8.011} under the ``Human Body – High Definition`` configuration , with internal camera calibration derived from the software’s native library. Feature matching and densification were executed, respectively, via the proprietary Samantha and Stasia algorithms, both implemented within a Multiview Stereo framework. This process yielded an initial sparse point cloud of 28,473 points and a dense cloud comprising 1,873,142 points, which was finally validated in MeshLab 2022.02 for both geometric accuracy and metric consistency. %The resulting datasets were archived on an external storage device managed by the scientific coordinator and integrated into the Omeka-S semantic database to facilitate automated documentation of process metadata.

\section{Dataset Creation}
\subsection{3D Model Preparation and Scene Setup}
\begin{figure}[htb]
\centering
\begin{minipage}[b]{0.2\linewidth}
  \centering
    \includegraphics[width=\linewidth]{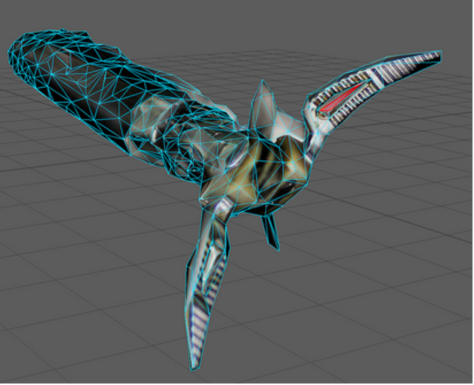} 
%  \vspace{1.5cm}
  \subcaption[first caption.]{}
  \label{fig:modeling_import}
\medskip
\end{minipage}
\begin{minipage}[b]{0.2\linewidth}
  \centering
    \includegraphics[width=\linewidth]{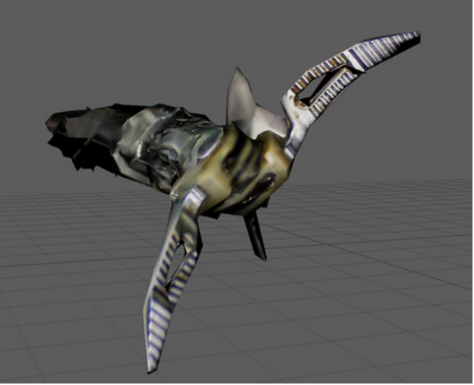}
%  \vspace{1.5cm}
  \subcaption[first caption.]{}
  \label{fig:modeling_holes}\medskip
\end{minipage}
\begin{minipage}[b]{0.2\linewidth}
  \centering
    \includegraphics[width=\linewidth]{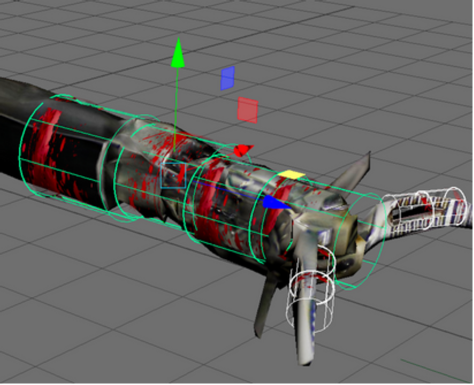}
%  \vspace{1.5cm}
  \subcaption[first caption.]{}
  \label{fig:modeling_blood}\medskip
  \end{minipage}
\begin{minipage}[b]{0.2\linewidth}
  \centering
    \includegraphics[width=\linewidth]{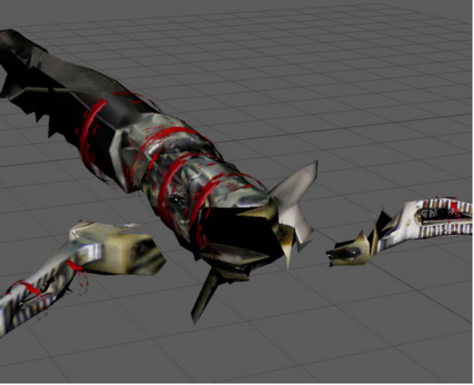}
%  \vspace{1.5cm}
  \subcaption[first caption.]{}
  \label{fig:modeling_anim}\medskip
  \end{minipage}
\caption{Synthetic modeling milestone: \ref{fig:modeling_import} import model obtained from photogrammetry, \ref{fig:modeling_holes} manual model refinement, \ref{fig:modeling_blood} add blood patches, \ref{fig:modeling_anim} divide the model into elements to be animated.}
\label{fig:mdeling}
\end{figure}

A 3D model of the robotic arms created using photogrammetry was imported and optimised in Autodesk Maya\texttrademark \  with a total of 1,882 vertices in 3,416 faces. In Fig.~\ref{fig:modeling_import}, we report an example. The mesh was cleaned up through topological refinement, the central body was anchored to a cylinder, and all vertices were moved to fit it. The terminal tools, such as clamps, joints, and tool elements, were finished by hand. The vertices were adjusted to reproduce the topological shape correctly, and finally, the model was perforated to accurately replicate the holes in the clamps, as shown in Fig. \ref{fig:modeling_holes}. 
The texture of the objects was applied from the beginning to preserve a realistic consistency. To simulate intraoperative visual conditions, variable reflectivities were assigned to surface materials, and random blood-like patches were added as independent texture layers. The blood patches were modelled as cylindrical meshes wrapped around the instruments and associated with random texture maps, providing realistic variability in appearance as shown in Fig. \ref{fig:modeling_blood}.
Each arm was divided into functional units with precisely positioned rotation points to ensure physically realistic rotations and translations. The vertices were remodeled so that there were no holes during all possible rotations of the units. The maximum rotation angle was taken empirically from the physical object as shown in Fig. \ref{fig:modeling_anim}.
%Figure~\ref{fig:modeling} illustrates the overall model result for each step of the pipeline.

% Below is an example of how to insert images. Delete the ``\vspace'' line,
% uncomment the preceding line ``\centerline...'' and replace ``imageX.ps''
% with a suitable PostScript file name.
% -------------------------------------------------------------------------
\begin{comment}
\begin{figure}[htb]
\begin{minipage}[b]{1.0\linewidth}
  \centering
  \centerline{\includegraphics[width=8.5cm]{img/Pulizia braccia.png}}
%  \vspace{2.0cm}
  \caption{Synthetic model modeling: from 3D model extraction to final result. \mg{non si capisce quali fasi sono presentate in figura, migliore descrizione nel testo e/o caption}}
  \label{fig:modeling}
\end{minipage}
\end{figure}
%
\end{comment}

\subsection{Automated Animation Pipeline}
A Python script has been developed to control the animation of three robotic arms, two clamps, and one scissor of the Da Vinci\texttrademark \  surgical robot. The script runs in Maya's standalone mode, allowing batch generation without user interaction. Key functionalities include:
\begin{comment}
Non mi piace usare il subsubsection, ma l'item mi indenta, da rivedere esteticamente
Non puoi cambiare lo stile della conferenza ... se non ti piace enumerate fai \subsubsection
\end{comment}
\begin{enumerate}
    \item Importing objects to be animated and displaying them in the scene. The number of objects and their initial position can be chosen randomly.
    \item Generation of random condition. Blood patches are applied around all objects presented in the scene and light can change the intensity within a range. 
    \item Tool trajectory randomization.
    %\end{enumerate}
    From the seed and the hash of the arm are generated two random sinusoidal functions to simulate the tool trajectories; the first is needed to open and close the tool smoothly, and the other to create the deterministic animations of each arm. All animations can oscillate within manually defined limits. It is possible to add a static offset to adjust the starting position and avoid overlaps, which can be generated randomly or set manually.

    %\begin{enumerate}
    %    \setcounter{enumi}{3}
    \item Green screen background generation for compositing.
    The green screen background is generated using a large-radius sphere (1000 units) covered with uniform green material. This technique enables the export of a video that can be easily superimposed on a real background. Green was chosen because it was the least prevalent color within the spectrum of the same image on a transparent background.

    %   \begin{enumerate}
    %   \setcounter{enumi}{4}
    \item Automatic playlist rendering to video. A preview was rendered directly from the viewport, in H.264 format, with a resolution of 1920x1080 and without graphic overlays. 
\end{enumerate}

\begin{figure}[h]
    \centering
    \includegraphics[width=0.8\columnwidth]{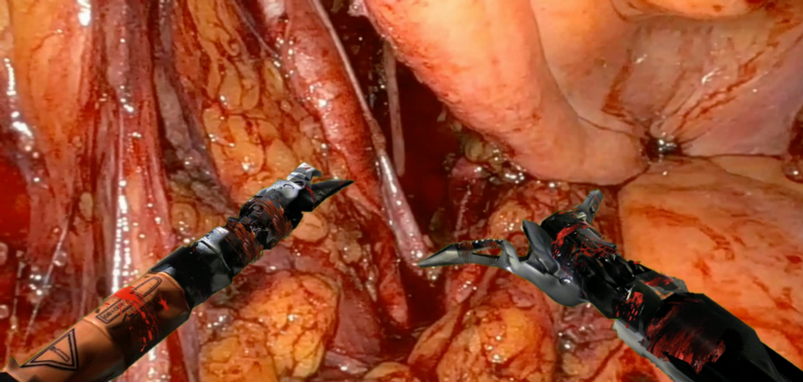}
    \caption{Frame extracted from synthetic video automatically generated. The robotic arms are animated on endoscopic background.}
    \label{fig:syntetic_example}
\end{figure}

%The fixed seed allows the reproducibility of the experiment but gives you the same videos in output. For this reason, you can set the seed but you should change it to obtain different videos.
The script has been designed to be versatile and modular, so that it can easily manage the presence of one, two or all mechanical arms simultaneously. At the same time, it maintains deterministic behaviour, thanks to the use of a seed that controls both the random selection of materials and the generation of animation curves, thus ensuring consistent and replicable results.
Each simulation cycle generates short clips with unique motion and lighting configurations, enabling the collection of large-scale synthetic data.
The last step is to replace the greenscreen with a real video that has been specifically recorded, at the end of a surgical operation  and without the insertion of robotic tools. %This allows for perfect consistency with the operating field and the overlaying of the robotic arms that were specifically generated. The background was replaced by recognising every pixel that was not green and replacing it pixel by pixel with the value of the endoscopic video. 
In Fig. \ref{fig:syntetic_example}, we can see an example of a frame produced by the pipeline. %The recognition was then used as a binary segmentation mask of the arms for each frame of the video.  
The final dataset includes RGB frames and corresponding binary segmentation masks of the robotic arms (these masks were generated automatically by rendering object IDs in Maya, providing pixel-level accuracy). Each synthetic scene was designed to replicate real intraoperative perspectives, with varying angles, instrument poses, and lighting conditions.

\section{Dataset validation}
To assess the efficiency and reliability of the constructed dataset, a targeted experiment was designed. This experiment involves training a deep learning model using various combinations of synthetic and real data. The primary goal of this analysis is to assess the impact of incorporating artificially generated data on the training of deep models and their potential to improve generalization. To validate our analysis, we utilize a real dataset in which robotic arms were manually segmented from actual videos. 

\subsection{Real Dataset}
The real dataset used in this study comprises approximately 1,000 images  with resolution $1920\x1080$, acquired in real surgical scenarios, directly obtained from the Da Vinci camera probe. %The images represent frames extracted from a video of a real surgical procedure and therefore constitute authentic data from actual procedures. 
All images were annotated manually using the Label Studio platform \cite{labelstudio}. %, an open-source software for managing and labelling datasets for machine learning tasks. 
%In particular, the "Segmentation with Mask" label was used, which allows the areas of interest within the images to be precisely delimited using segmentation masks. The frame were labelled manually one by one, 
% an example is shown in Fig. \ref{fig:real_withMask}.}
We fixed $80\%$ of the video frames for the training set, $10\%$ for the validation set, and the remaining $10\%$ for the test set. To avoid leakage between training and testing, consecutive frames from the same video cannot be included in different sets.
%In the train-val-test split, 
%We avoid the possibility that close and therefore very similar frames would end up in different datasets, otherwise there would be contamination. To perform this, we extracted all the frames from the videos, maintaining the chronological order, and then separated the frames into the corresponding datasets with 2 cuts. Only at this point a shuffle was performed within the individual datasets.

\subsection{Training dataset}
The training dataset is created dynamically by adding a variable fraction of synthetic images $\alpha = \frac{n_{\text{synt}}}{n_{\text{tot}} }$, with $n_{\text{tot}}=n_{\text{real}}+ n_{\text{synt}}$ where $n_{\text{synt/real}}$ is the number of synthetic and real images, respectively. The goal is to analyze how the presence of synthetic content affects the model's ability to segment on the real test dataset. Both the validation and test datasets remain 100\% real. 

\subsection{Training of segmentation models}
Dataset validation was performed training a UNet \cite{unet} architecture using several backbones,  namely ResNet18 \cite{resnet18}, VGG \cite{vgg16}, and ResNeXt-50 32x4d \cite{resnext}. %to provide as general an evaluation as possible. 
%The pipeline was implemented in a Jupyter Notebook environment \cite{Jupyter_Notebooks} and We used PyTorch Lightning \cite{pytorch_lightning} for training and Segmentation Models PyTorch (SMP)\cite{Segmentation_Models_Pytorch} for building the segmentation model.
%\subsubsection{Evaluation metrics: Dice score, Dice loss and IoU}
Training was conducted using the Adam optimizer with a constant learning rate $10^{-4}$ and Dice loss. Each model was trained for a maximum of 20 epochs, %at which point the performance computed on the validation set reached a plateau. It is also used 
with early stopping based on the performance on the validation set. As validation metrics, we used the Dice score. %which is the value obtained by dividing the double of the intersection between the predicted mask and the real mask by the sum of the predicted and real masks, as shown in equation \eqref{Dice score}. As for 
%The loss metric, we used the Dice loss, simply described by equation $Loss=1-Dice$ as complement to 1 of the Dice score, so that minimizing the loss results in better performance. 
Performances were evaluated on the test dataset using Intersection over Union (IoU). %coefficient metrics described in the equation \eqref{IoU}.
\begin{comment}
\begin{minipage}{0.94\linewidth}
    \begin{minipage}{0.44\linewidth}
        \begin{equation}
            \text{Dice} = \frac{2 \cdot |A \cap B|}{|A| + |B|} \label{Dice score}
        \end{equation}
    \end{minipage}
    \hfill
    \begin{minipage}{0.48\linewidth}
        \begin{equation}
            \text{IoU}_{(A, B)} = \frac{|A \cap B|}{|A \cup B|}
            \label{IoU}
        \end{equation}
    \end{minipage}
\end{minipage}
%\vspace{.5em}
\end{comment}
\begin{comment}
    Integration with Weights \& Biases \cite{wandb} allows you to inspect loss and performance curves retrospectively and compare different runs with each other.
\end{comment}
%\subsubsection*{Augmentaiton}
%To complete the overview of the data, we added augmentation.
Data augmentation was used to improve generalization. %The augmentation strategy balances geometric and photometric transformations with the aim of improving the model's generalization ability with respect to variations in pose, scale, lighting, and noise.
The training uses a set of transformations that includes horizontal flips, a controlled combination of translations and scale changes without rotation to maintain as much realism as possible, and minimal padding followed by random cropping to a resolution of $256 \x 256$. 
The geometric changes are accompanied by photometric perturbations such as Gaussian noise, perspective distortions, and variations in contrast, brightness, gamma, hue, and saturation, organized into mutually exclusive groups to avoid over-transformation in a single sample. 
%Although the dataset is synthetic, the use of data augmentation is still useful. The generation of the dataset produces new images, but augmentation introduces additional controlled variations that are too expensive to include in the generation of the dataset.

\section{Experimental Results}
\begin{figure}[h]

\begin{minipage}[b]{1.0\linewidth}
  \centering
  \includegraphics[width=\linewidth]{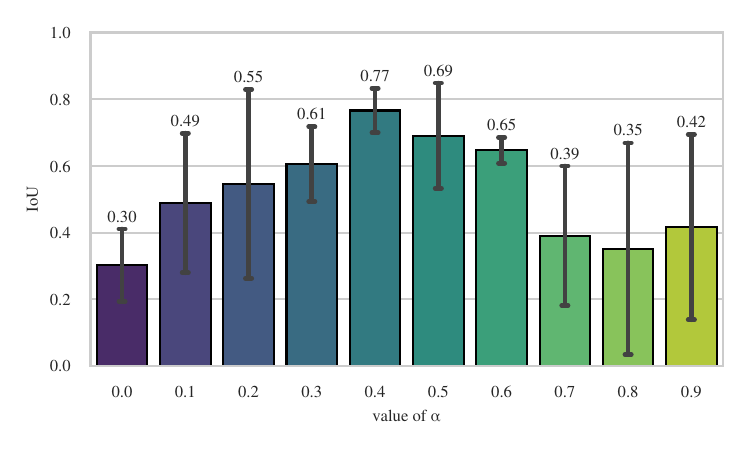}
  
%  \vspace{2.0cm}
  \subcaption[]{Mean of performance over models.}
  \label{fig:validation_meaning}\medskip
\end{minipage}
\begin{minipage}[b]{.32\linewidth}
  \centering
\includegraphics[width=\linewidth]{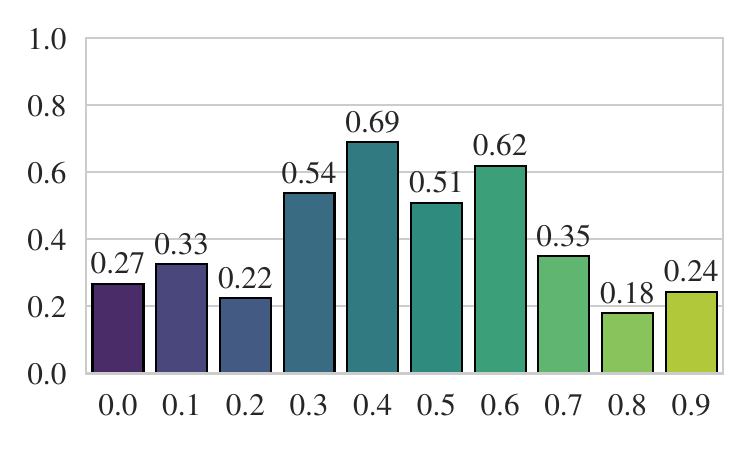}
%  \vspace{1.5cm}
  \subcaption[]{ResNet18}
  \label{fig:validation_resnet18}\medskip
  \end{minipage}
\hfill
\begin{minipage}[b]{0.32\linewidth}
  \centering
\includegraphics[width=\linewidth]{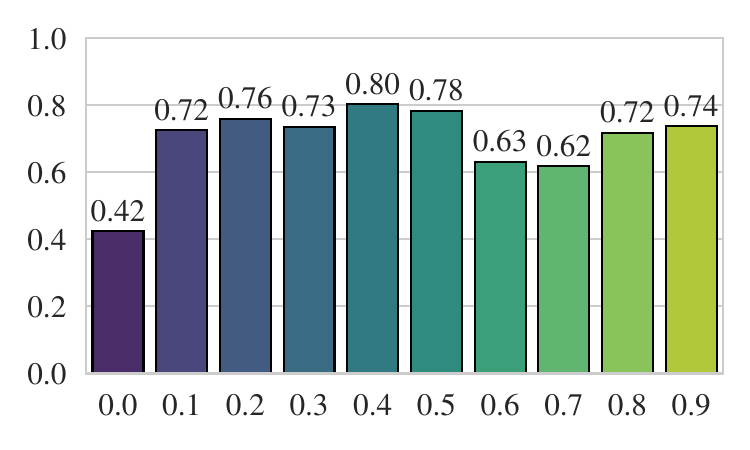}
%  \vspace{1.5cm}
  \subcaption[]{VGG16}
  \label{fig:validation_vgg16}\medskip
  \end{minipage}
\hfill
\begin{minipage}[b]{0.32\linewidth}
  \centering
\includegraphics[width=\linewidth]{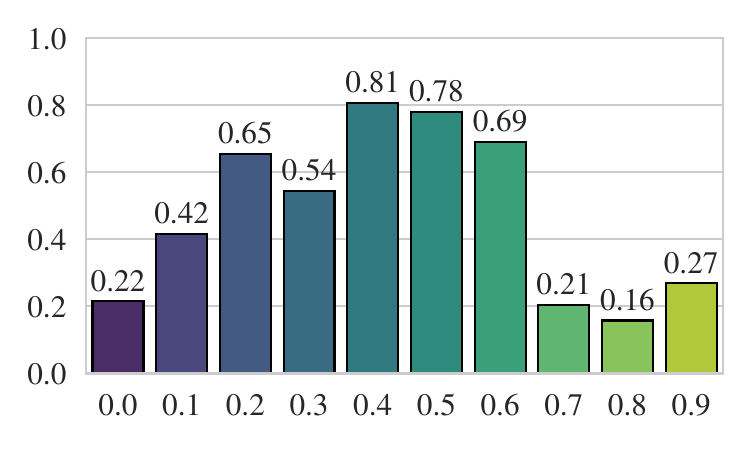}
%  \vspace{1.5cm}
  \subcaption[]{ResNeXt-50}
  \label{fig:validation_resnext}\medskip
\end{minipage}
\caption{Quantitative comparison of IoU performance in relation to $\alpha$. In \ref{fig:validation_meaning} Each column rappresent the mean and confidence interval of the IoU performance obtained over all models. From \cref{fig:validation_resnet18,fig:validation_vgg16,fig:validation_resnext} are represented the IoU performances for each model. }
\label{fig:validazione_grafico}
\end{figure}
\begin{figure}[h]
  \centering
  \includegraphics[width=0.8\linewidth]{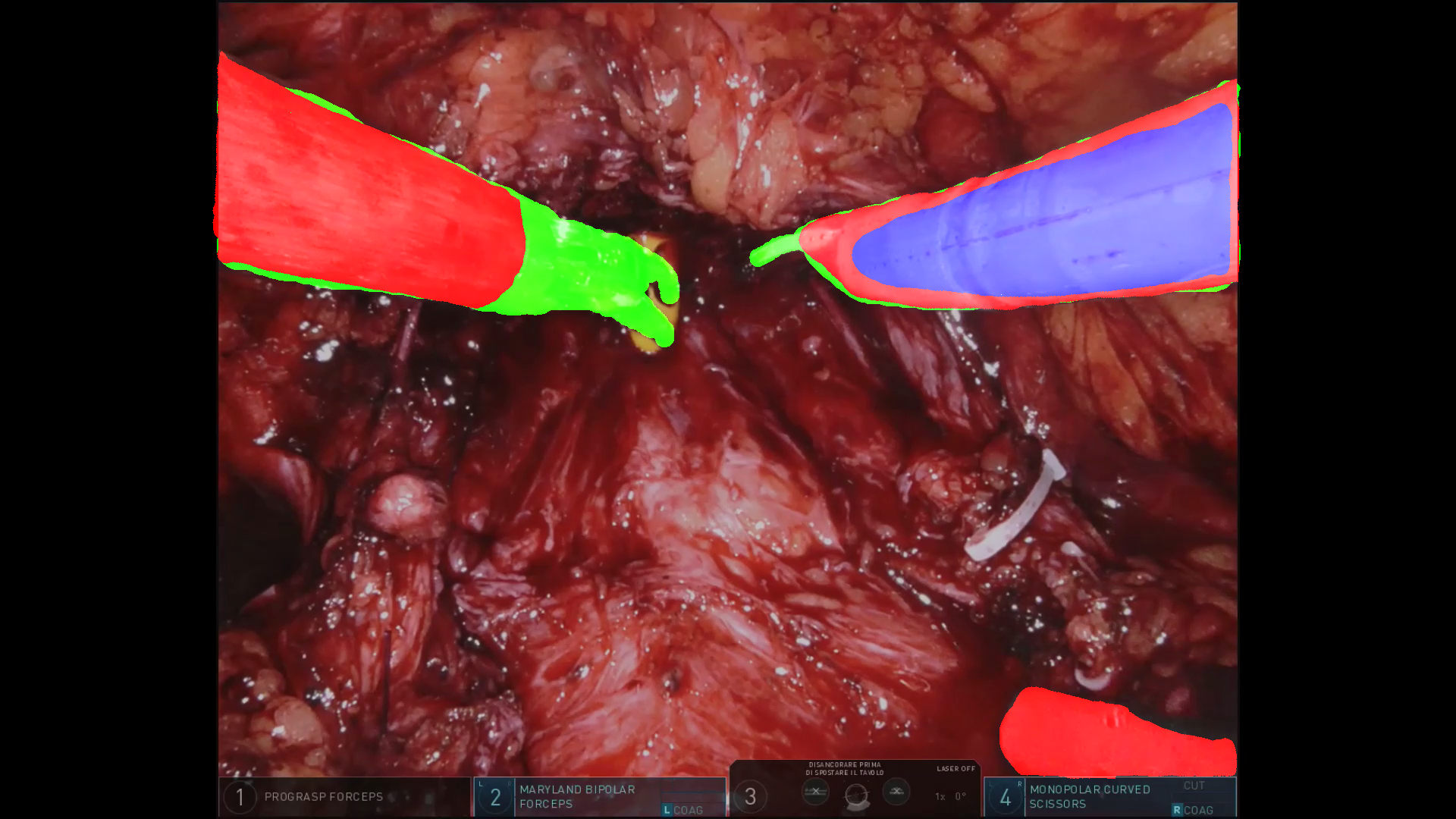} % 8.5cm
  \caption{Example of real frame and corresponding \textcolor{green}{ground truth} in  \textcolor{green}{green}. In  \textcolor{blue}{blue} and in  \textcolor{red}{red} the segmentation of the VGG16 model trained on  \textcolor{blue}{$\alpha=0$} (only real data) and  \textcolor{red}{$\alpha = 0.4$} respectively.} \medskip
  \label{fig:real_withMask}
\end{figure}

%Fig. \ref{fig:validazione_grafico} shows a bar plot of the IoU obtained from sequential execution of the pipeline for each backbone model, where the percentage between synthetic and real data included in the training set  ($\alpha$) is increased by 10\% for each run. 
In Fig.~\ref{fig:validation_meaning} shows the mean performance in terms of IoU (and the corresponding confidence interval) obtained by all the tested backbones as a function of $\alpha$ . % we include also the confidence interval guided from standard deviation. 
In Figs. \ref{fig:validation_resnet18}, \ref{fig:validation_vgg16}, \ref{fig:validation_resnext} are detailed the performance of single backbones. 
These results indicate that models trained with a balanced mixture of real and synthetic data achieved the highest segmentation performance, outperforming the configuration that used only real data. In contrast, an excess of synthetic data may introduce a significant domain shift, which reduces generalization to real test samples. Notably, the low IoU values below $0.4$ obtained with our small-scale real dataset can be increased to approximately $0.8$ by adding a 50\% fraction of synthetic images in the training set. In Fig. \ref{fig:real_withMask} is shown the qualitative improvement of the segmentation when the model VGG16 is trained on $\alpha=0$,in blue, and $\alpha=0.4$,in red. The manually labeled ground truth is represented in green.

\begin{comment}    
\begin{table}[ht]
\centering
\caption{Segmentation results under different traning models for each dataset composition.}
\label{tab:tabella_risultati_modelli}
\begin{tabular}{lcccccccccc}
\toprule
\textbf{Models} & \textbf{0\%} & \textbf{20\%} &  \textbf{40\%}  & \textbf{60\%}   & \textbf{80\%}  & \textbf{99\%} \\
\midrule
resnet 18       & -  & -  &   -    &   -    &    -   &     -  \\
resnet 34       & -  & -  &    -   &   -    &   -    &   -    \\
resnet 50       & -  & -  &   -    &   -    &   -    &   -    \\
vgg 16          &   -    &  -     &   -    &  -     &  -     &  -     \\
densenet121     &  -     &     -  &   -    &  -     &   -    &  -     \\
\bottomrule
\end{tabular}
\end{table}

\end{comment}

% \begin{comment}
    
% \begin{figure}[htb]
% \begin{minipage}[b]{1.0\linewidth}
%   \centering
%   \centerline{\includegraphics[width=8.5cm]{img/grafico_validazione.png}}
% %  \vspace{2.0cm}
%   \caption{Qualitative comparison of segmentation masks from real-only to mixed dataset. Each column rappresent the mean of the performance obtained from all model.}\medskip
%   \label{fig:validazione_grafico}
% \end{minipage}
% \end{figure}

% \end{comment}

\section{Discussion}
Our results highlight two key findings. First, adding synthetic data to the dataset enhances its geometric and photometric diversity, thereby helping models generalize more effectively. However, excessively relying on synthetic data can reduce the model’s alignment with real-world data, leading to a decline in performance. This phenomenon, known as Domain Shift \cite{Domain_Shift}, is a fundamental issue in machine learning and domain adaptation. Here, models trained on a source distribution often perform poorly on a different, target distribution. Such performance drops can be triggered by a range of intrinsic factors, including differences in lighting, angles, colors, or textures. Because artificial vision models are sensitive to even these minor variations, Domain Shift becomes a critical challenge. 

\section{Conclusion}

We presented a comprehensive framework for creating and validating a synthetic dataset for the segmentation of robotic surgery instruments. The automated simulation pipeline generates photorealistic, diverse, and reproducible data suitable for training deep segmentation models. Experimental validation showed that a balanced combination of synthetic and real data improves model performance, supporting the integration of synthetic datasets as a strategic component of surgical AI research. 
% Access to the images is restricted to protect sensitive information from the surgical procedures. 
% Consequently, only the synthetic data and pipeline can be published on \url{https://github.com/EIDOSLAB/Sintetic-dataset-DaVinci}. 
Currently, only three surgical instruments have been reconstructed, but the pipeline is parametric and can be extended to support novel instruments and tasks. Future work can also involve training solely on a synthetic dataset and then fine-tuning on real data.  This combination may result in more robust, adaptive, and accurate models, exploiting the best aspects of both datasets: the scalability and flexibility of synthetic data and the realism of actual data. 

% \subsection{Limitation}

% We can conclude that the proposed synthetic dataset appears to be efficient in improving model performance, but its optimal impact must be verified on a case-by-case approach. It is recommended to conduct several tests to verify the percentage of synthetic datasets to be applied in each specific case.

% Although we have tried to maintain as much realism as possible, this could be improved, possibly by replacing random variables with elements having real distributions, such as reproductions of real actions.

% Currently, only three surgical instruments have been reconstructed, but the pipeline is parametric and remains unchanged with the introduction of different instruments.

% % --------------------------------------------------------------------------
% % ETHICAL COMPLIANCE
% % --------------------------------------------------------------------------
\noindent\textbf{Compliance with Ethical Standards}
This study was performed in accordance with the Declaration of Helsinki and approved by the institutional ethics committee. All patient data were anonymized prior to processing.
%\section{Data Availability}
%\gio{Qui o nell'abstract ?}
%The real dataset and the robotic arm model is published on huggingface on \url{https://huggingface.co/datasets/rossbina/tesi-dataset}.
%The syntetic generation pipeline is published on \url{https://github.com/EIDOSLAB/Sintetic-dataset-DaVinci}

%\section{Acknowledgment}
%This work was supported by the Ospedale San Luigi of Turin and the Department of Computer Science, University of Turin.

\bibliographystyle{IEEEbib}
\bibliography{refs}

\end{document}